\documentclass[11pt,a4paper]{article}
\usepackage[hyperref]{acl2020}
\usepackage{times}
\usepackage{latexsym}

\usepackage{natbib}
\usepackage[utf8]{inputenc}
\usepackage{tabularx,graphicx}
\usepackage{xcolor}
\usepackage{subcaption}
\usepackage{booktabs} 
\usepackage{todonotes}

\aclfinalcopy

\title{A Visuospatial Dataset for Naturalistic Verb Learning}

\author{Dylan Ebert \\
    Brown University \\
    dylan\_ebert@brown.edu \\\And
    Ellie Pavlick \\
    Brown University \\
    ellie\_pavlick@brown.edu \\
}

\date{}

\begin{document}

\maketitle

\begin{abstract}
    
We introduce a new dataset for training and evaluating grounded language models. Our data is collected within a virtual reality environment and is designed to emulate the quality of language data to which a pre-verbal child is likely to have access: That is, naturalistic, spontaneous speech paired with richly grounded visuospatial context. We use the collected data to compare several distributional semantics models for verb learning.
We evaluate neural models based on 2D (pixel) features as well as feature-engineered models based on 3D (symbolic, spatial) features, and show that neither modeling approach achieves satisfactory performance. Our results are consistent with evidence from child language acquisition that emphasizes the difficulty of learning verbs from naive distributional data. We discuss avenues for future work on cognitively-inspired grounded language learning, and release our corpus with the intent of facilitating research on the topic.

 
    
\end{abstract}

\section{Introduction}

While distributional models of semantics have seen incredible success in recent years \cite{devlin2018bert}, most current models lack ``grounding'', or a connection between words and their referents in the non-linguistic world. Grounding is an important aspect to representations of meaning and arguably lies at the core of language ``understanding'' \cite{bender-koller-2020-climbing}. Work on grounded language learning has tended to make opportunistic use of large available corpora, e.g. by learning from web-scale corpora of image \cite{bruni2012distributional} or video captions \cite{sun2019videobert}, or has been driven by particular downstream applications such as robot navigation \cite{mattersim}. 

In this work, we take an aspirational look at grounded distributional semantics models, based on the type of situated contexts and weak supervision from which children are able to learn much of their early vocabulary. Our approach is motivated by the assumption that building computational models which emulate human language processing is in itself a worthwhile endeavor, which can yield both scientific \cite{potts2019case} and engineering \cite{linzen-2020-accelerate} advances in NLP. Thus, we aim to develop a dataset that better reflects both the advantages and the challenges of humans' naturalistic learning environments. For example, unlike most vision-and-language models, children likely have the advantage of access to symbolic representations of objects and their physics prior to beginning word learning \cite{spelke2007core}. However, also unlike NLP models, which are typically trained on image or video captions with strong signal, children's language input is highly unstructured and the content is often hard to predict given only the grounded context \cite{gillette1999human}. 


We make two main contributions. First (\S\ref{sec:data}), using a virtual reality kitchen environment, we collect and release\footnote{\url{https://github.com/dylanebert/nbc}} the New Brown Corpus\footnote{Our university namesake, plus paying homage to important Brown corpora in both NLP \cite{francis1979brown} and Child Language Acquisition \cite{brown1973first}.}: A dataset containing 18K words of spontaneous speech alongside rich visual and spatial information about the context in which the language occurs. Our protocol is designed to solicit naturalistic speech and to have good coverage of vocabulary items with low average ages of acquisition according to data on child language development \cite{frank2017wordbank}. 
Second (\S\ref{sec:exp-design}), we use our corpus to compare several distributional semantics models, specifically comparing models which represent the environment in terms of objects and their physics to models which represent the environment in terms of pixels. We focus on verbs, which have received considerably less attention in work on grounded language learning than have nouns and adjectives \cite{forbes-etal-2019-neural}. More so than nouns, verb learning is believed to rely on subtle combinations of both syntactic and grounded contextual signals \cite{piccin2007nouns} and thus progress on verb learning is likely to require new approaches to modeling and supervision. In our experiments, we find that strong baseline models, both feature-engineered and neural network models, perform only marginally above chance. However, comparing models reveals intuitive differences in error patterns, and points to directions for future research.

\section{Data}
\label{sec:data}

The goal of our data collection is to enable research on grounded distributional semantics models using data that better resembles the type of input young children receive on a regular basis during language development. Doing this fully is ambitious if not impossible. Thus, we focus on a few aspects of children's language learning environment that are lacking from typical grounded language datasets and that can be emulated well given current technology: 1) spontaneous speech (i.e. as opposed to contrived image or video captions) and 2) rich information about the 3D world (i.e. physical models of the environment as opposed to flat pixels).

We develop a virtual reality (VR) environment within which we collect this data in a controlled way. Our environment data is described in Section \ref{sec:env} and our language data is described in Section \ref{sec:protocol}. Our collection process results in a corpus of 152 minutes of concurrent video, audio, and ground-truth environment information, totaling 18K words
across 18 unique speakers performing six distinct tasks each. The current data is available for download in json format at \url{https://github.com/dylanebert/nbc}. The code needed to implement the described environment and data recording is available at \url{https://github.com/dylanebert/nbc_unity_scripts}.

\subsection{Environment Data Collection}
\label{sec:env}

\subsubsection{Environment Construction}

Our environment is a simple kitchen environment, implemented in Unity with SteamVR and our experiments are conducted using an HTC Vive headset. We choose to use VR as opposed to alternative interfaces for simulated interactions (e.g.\ keyboard or mouse control) since VR enables participants to use their usual hand and arm motions and to narrate in real time, leading to more natural speech and more faithful simulations of the actions they are asked to perform.

\begin{figure}[ht!]
\centering
	\begin{subfigure}{0.45\linewidth} 
		\includegraphics[width=\textwidth]{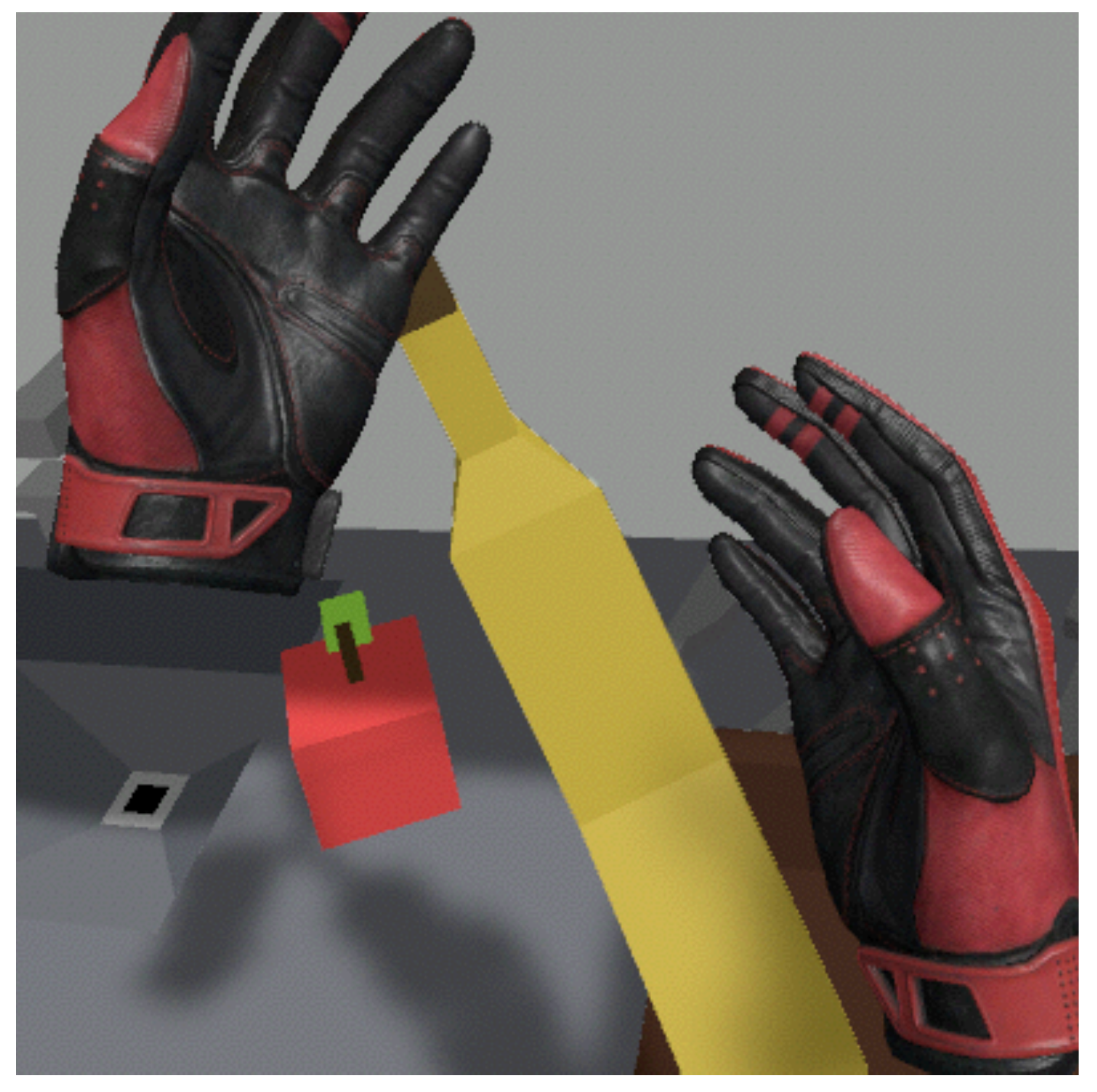} 
	\end{subfigure} %
	\begin{subfigure}{0.45\linewidth} 
		\includegraphics[width=\textwidth]{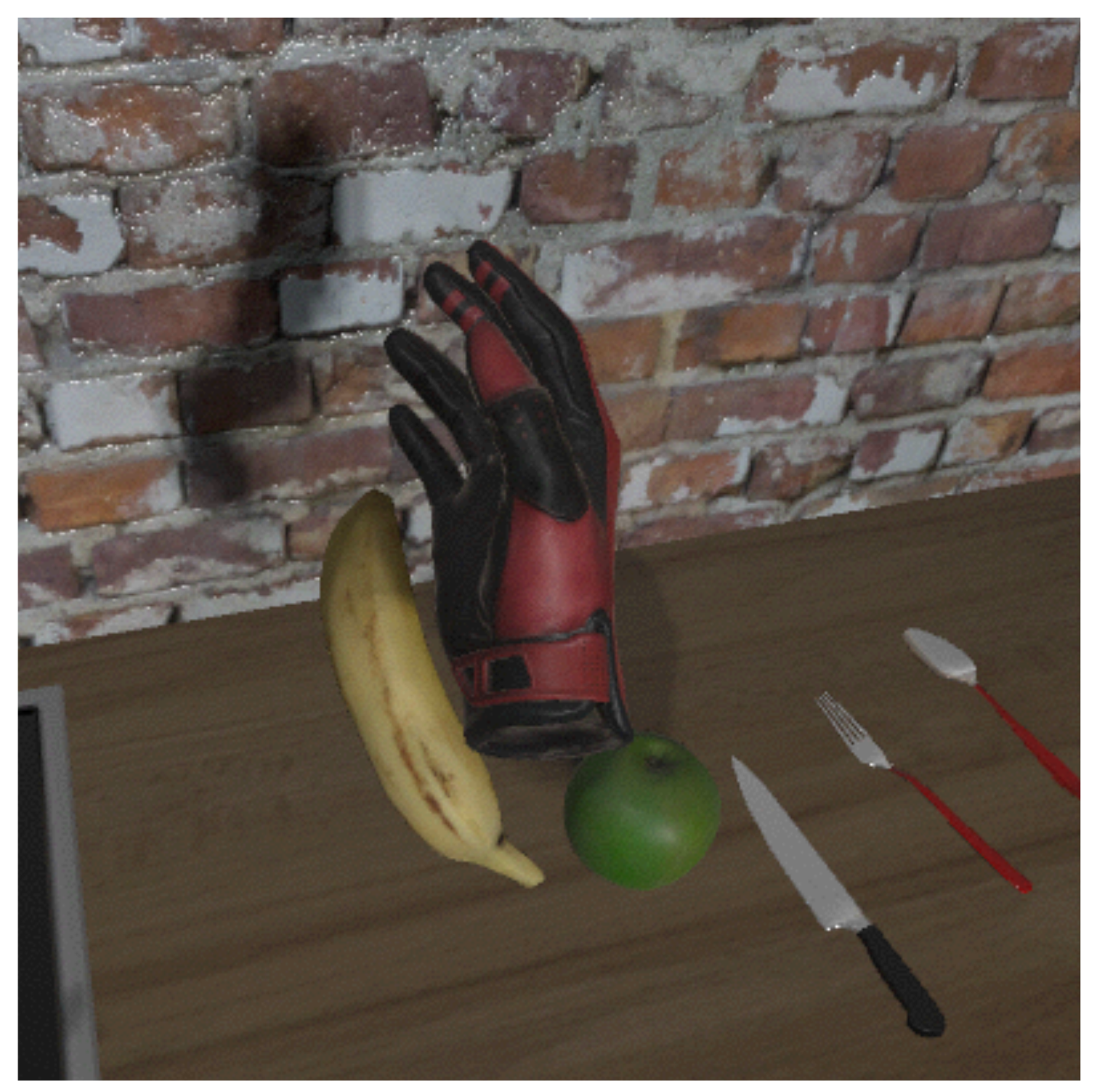}
	\end{subfigure}
	\caption{Screenshots of a person picking up a banana in each of our two kitchen aesthetics.} 
\label{fig:block-v-real}
\end{figure}

We design six different kitchen environments, using two different visual aesthetics (Fig.\ \ref{fig:block-v-real}) with three floorplans each. This variation is so that we can test, for example, that learned representations are not overfit to specific pixel configurations or to exact hand positions that are dependent on the training environment(s) (e.g.\ ``being in the northwest corner of the kitchen'' as opposed to ``being near the sink''). Each kitchen contains at least 20 common objects (not every kitchen contains every object). These objects were selected because they represent words with low average ages of acquisition (described in detail in \S\ref{sec:protocol}) and were available in different Unity packages and thus could be included in the environment with different appearances. Across all kitchens, the movable objects used are: \texttt{Apple}, \texttt{Ball}, \texttt{Banana}, \texttt{Book}, \texttt{Bowl}, \texttt{Cup}, \texttt{Fork}, \texttt{Knife}, \texttt{Lamp}, \texttt{Plant}, \texttt{Spoon}, \texttt{Toy1:Bear$\mid$Bunny}, \texttt{Toy2:Doll$\mid$Dinosaur}, \texttt{Toy3:Truck$\mid$Plane}. The participant's hands and head are also included as movable objects. We also include the following immovable objects: \texttt{Cabinets}, \texttt{Ceiling}, \texttt{Chair}, \texttt{Clock}, \texttt{Counter}, \texttt{Dishwasher}, \texttt{Door}, \texttt{Floor}, \texttt{Fridge}, \texttt{Microwave}, \texttt{Oven}, \texttt{Pillar}, \texttt{Rug}, \texttt{Sink}, \texttt{Stove}, \texttt{Table}, \texttt{Trash Bin}, \texttt{Wall}, \texttt{Window}. 

Our environments are constructed using a combination of Unity Asset Store assets and custom models. All paid assets (most objects we used) come from two packs: 3DEverything Kitchen Collection 2 and Synty Studios Simple House Interiors, from the Unity asset store\footnote{\url{https://assetstore.unity.com/}}. These packs account for the two visual styles. VR interaction is enabled using the SteamVR Unity plugin, available for free on the Unity asset store. 

\begin{table*}[ht!]
\small
\centering 
\begin{tabular}{lp{.8\linewidth}}
	\toprule
\bf Name (Type) & \bf Description \\
    \midrule
pos (xyz) & Absolute position of object center, computed using the \texttt{transform.position} property; equivalent to position relative to an arbitrary world origin, approximately in the center of the floor.\\
rot (xyzw) & Absolute rotation of object, computed using the \texttt{transform.rotation} property.\\
vel (xyz) & Absolute velocity of object center, computed using the \texttt{VelocityEstimator} class included with SteamVR.\\
relPos (xyz) & Position of object's center relative to the person's head, computed using Unity's built-in \texttt{head.transform.TransformPoint(objectPosition)}.\\
relRot (xyzw) & Rotation of object relative to the person's head, computed by applying the inverse of the head rotation to the object rotation. \\
relVel (xyz) & Velocity of the object's center, from the frame of reference of the person's head \\
bound (xyz) & Distance from object's center to the edge of bounding box \\
inView (bool) & Whether or not the object was in the person's field of view, computed using Unity's \texttt{GeometryUtility} to compute if an object is in the Camera renderer bounds. This is based on the default camera's 60 degree FOV, not the wide headset FOV. The head and hands are always considered \textit{inView}.\\
img\_url (img) & Snapshot of the person's entire field of view as a 2D image. We compute this once per frame (as opposed to the above features which are computed once per object per frame). \\
    \bottomrule
\end{tabular}
\caption{Object features recorded during data collection. Object appearance does not vary across frames; img\_url does not vary across objects. All other features vary across object and frame.}
\label{tab:spatial-features}
\end{table*}

\subsubsection{Data Recording}

During data collection, we record the physical state of each object in the environment, according to the ground-truth in-game data, at a rate of 90fps (frames per second). The Vive provides accurate motion capture, allowing us to record the physical state of the user's head and hands \cite{borges2018htc} as well. For each object, we record the physical features described in Table \ref{tab:spatial-features}. Audio data is also collected in parallel to spatial data, using the built-in microphone. We later transcribe the audio using the Google Cloud Speech-to-Text API\footnote{\url{https://cloud.google.com/text-to-speech/}}. Word-level timestamps from the API allow us to match words to visuospatial frames. While spatial and audio data are recorded in real-time, video recording is not, since this would introduce high computational overhead and drop frames. Instead, we iterate back over the spatial data, and reconstruct/rerender the playback frame-by-frame. This approach makes it possible to render from any perspective if needed, though our provided image data is only from the original first-person perspective.

\subsection{Language Data Collection}
\label{sec:protocol}

We design our protocol so as to solicit the use of vocabulary items that are known to be common among children's early-acquired words. To do this, we first select 20 nouns, 20 verbs, and 20 prepositions/adjectives which have low average ages of acquisition according to \citet{frank2017wordbank} and which can be easily operationalized within our VR environment (e.g. \textit{``apple''}, \textit{``put (down)''}, \textit{``red''}, see Appendix \ref{app:vocab} for full word list). We then choose six basic tasks which the participants will be instructed to carry out within the environment. These tasks are: set the table, eat lunch, wash dishes, play with toys, describe a given object, and clean up toys.
The tasks are intended to solicit use of many of the target vocabulary items without explicitly instructing participants to use specific words, since we want to avoid coached or stilted speech as much as possible. One exception is the ``describe a given object'' task, in which we ask participants to describe specific objects as though a child has just asked what the object is, e.g.\ \textit{``What's a spoon?''}. We use this task to ensure uniform coverage of vocabulary items across environments, so that we can construct good train/test splits across differently appearing environments. See Appendix \ref{app:conditions} for details on distributing vocabulary items. 

We recruited 18 participants for our data collection. Participants were students and faculty members from multiple departments involved with language research. We asked each participant to perform each of our tasks, one by one, and to narrate their actions as they went, as though they were a parent or babysitter speaking to a young child. The exact instructions given to participants before each task are shown in Appendix \ref{app:script}. An illustrative example of the language in our corpus is the following: \textit{``okay let's pick up the ball and play with that will it bounce let's see if we can bounce it exactly let's let it drop off the edge yes it bounced the ball bounced pick it up again...''}. The full data can be browsed at \url{https://github.com/dylanebert/nbc}.

Our study design was determined not to be human subjects research by the university IRB. All participants were informed of the purpose of the study and provided signatures consenting to the recording and release of their anonymized data for research purposes (consent form in Appendix \ref{app:consent}). 

\subsection{Comparison to Child Directed Speech}

Since our stated goal was to collect data that better mirrors the distribution of language input a young child is likely to receive, we run several corpus analyses to assess whether this goal was met.\footnote{We focus our analysis only on the vocabulary composition of the transcribed data. Audio data is released, but the study of audio features (e.g.\ prosody, a key feature of CDS) is left for future work.} 

\subsubsection{Vocabulary Distribution}
\label{sec:cps:vocab}

First, we compare the distribution of vocabulary in our collected data to that observed in the Brent-Siskind Corpus \citep{brent2001role}, a corpus of child-directed speech consisting of 16 English-speaking mothers speaking to their pre-verbal children. For reference, we also compare with the vocabulary distributions of three existing corpora which could be used for training distributional semantics models:\ 1) MSR-VTT \citep{Xu:CVPR16}, a large dataset of YouTube videos labeled with captions, 2) Room2Room (R2R) \citep{mattersim}, a dataset for instruction following within a 3D virtual world, and 3) a random sample of sentences drawn from Wikipedia. Since our primary focus is on grounded language, MSR and R2R offer the more relevant points of comparison, since each contains language aligned with some kind of grounded semantic information (raw RGB video feed for MSR and video+structured navigation map for R2R). We include Wikipedia to exemplify the type of web corpora that are ubiquitous in work on representation learning for NLP. 

\begin{figure}[ht!]
\centering
	\begin{subfigure}{0.5\textwidth} 
		\includegraphics[width=\textwidth]{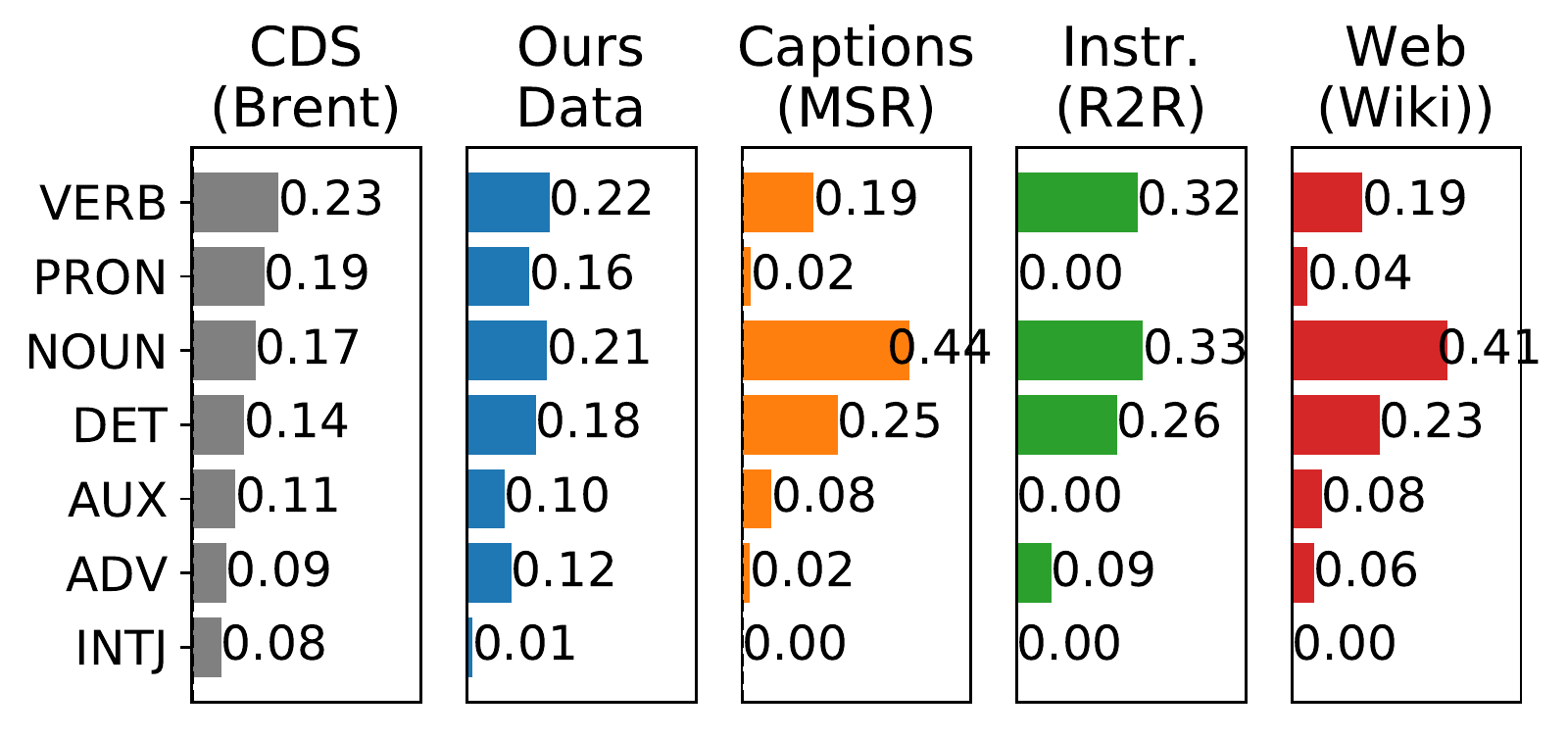} 
		\caption{Token-Level Frequency of Word Categories}
		\label{fig:tok-cat-dist}
	\end{subfigure} 
	\begin{subfigure}{0.5\textwidth} 
		\includegraphics[width=\linewidth]{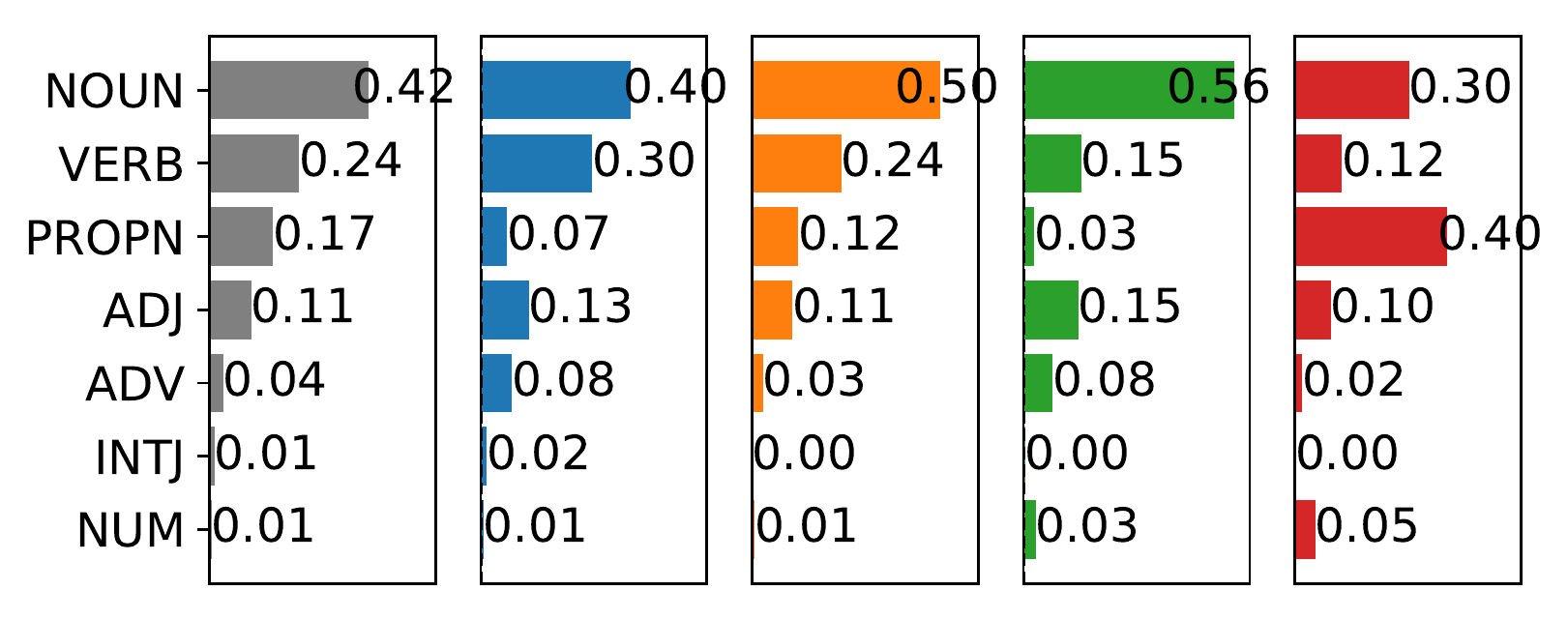}
		\caption{Type-Level Frequency of Word Categories}
		\label{fig:typ-cat-dist}
	\end{subfigure} 
	\begin{subfigure}{0.5\textwidth} 
		\includegraphics[width=\textwidth]{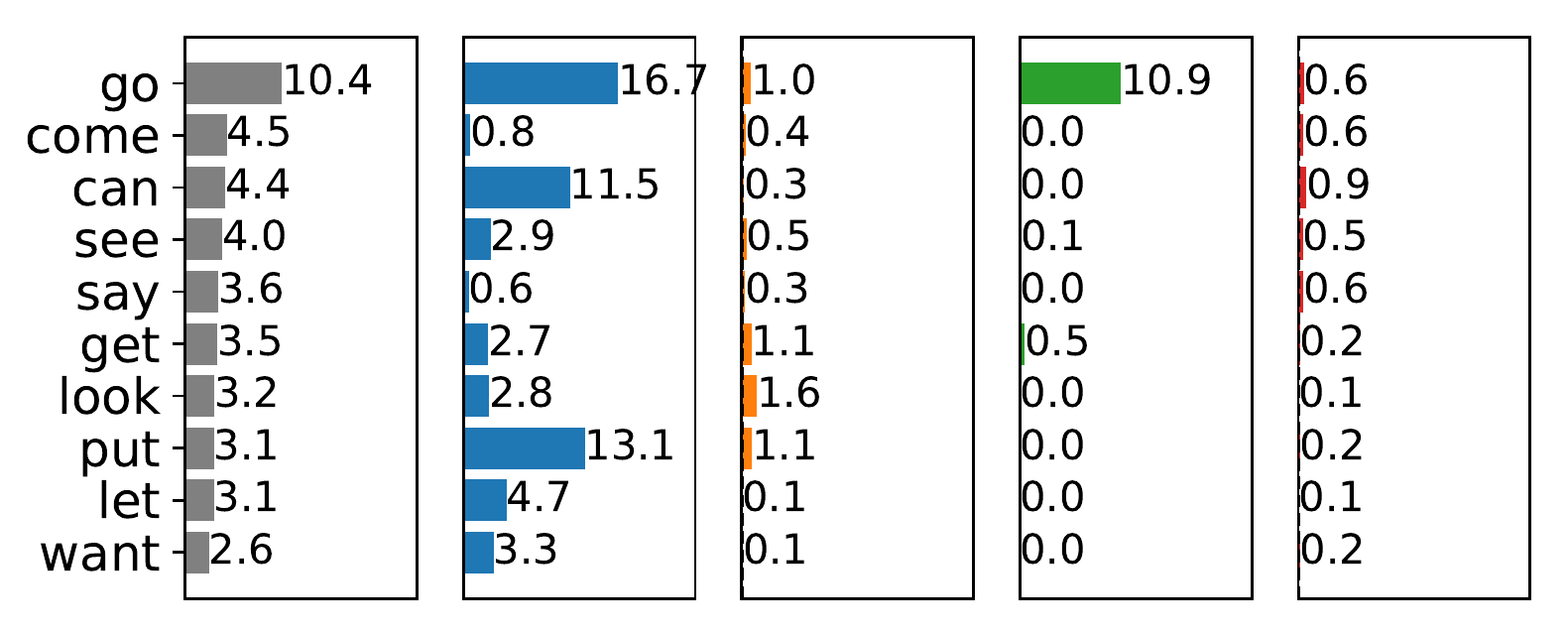}
		\caption{Token Frequency of Individual Verbs}
		 \label{fig:verb-dist}
	\end{subfigure}
	\begin{subfigure}{0.5\textwidth} 
		\includegraphics[width=\textwidth]{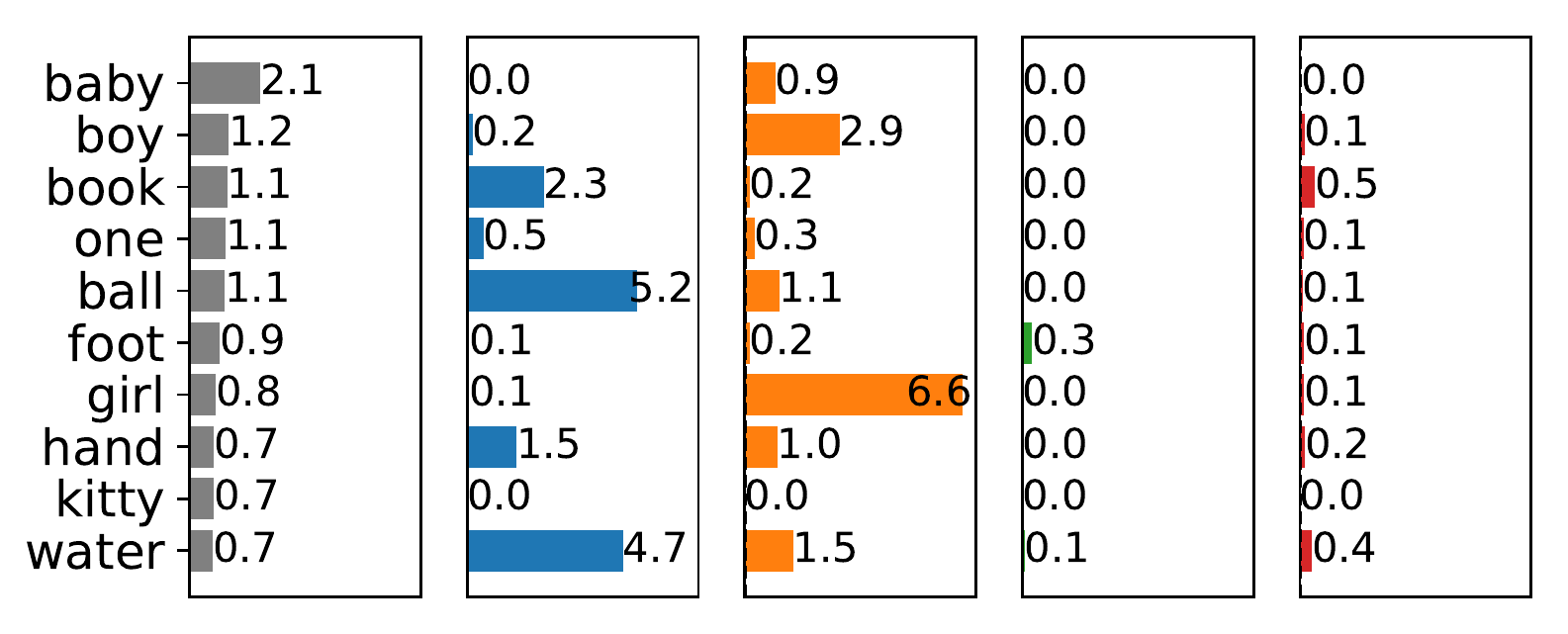}
		\caption{Token Frequency of Individual Nouns}
		 \label{fig:verb-dist}
	\end{subfigure}
	\begin{subfigure}{0.5\textwidth} 
		\includegraphics[width=\textwidth]{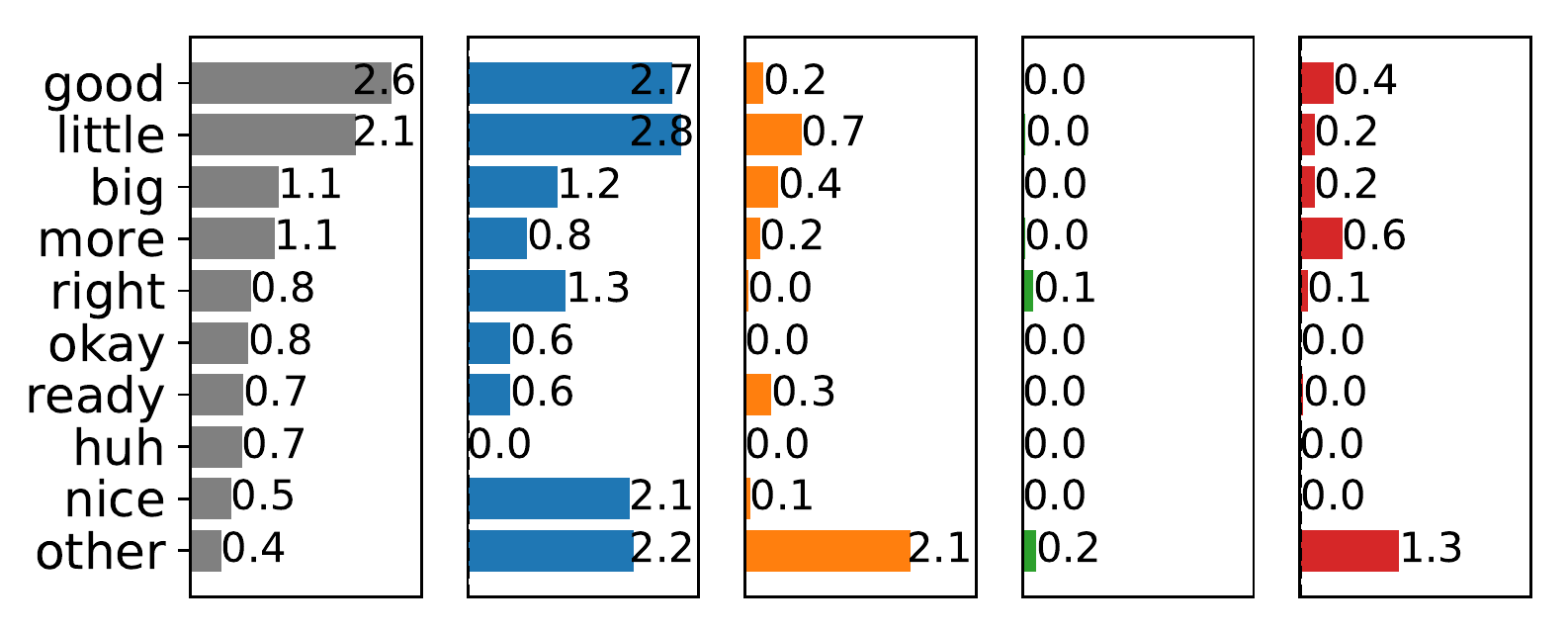}
		\caption{Token Frequency of Individual Adjectives}
		 \label{fig:adj-dist}
	\end{subfigure}
	\caption{Comparison of word category and lexical distributions. Lexical item frequencies labels are $\times1000$. Distributions are over the most frequent categories/words according to the Brent-Siskind corpus of child-directed speech.}
	\label{fig:vocab-dists}
\end{figure}

Figure \ref{fig:vocab-dists} shows, for each of the five corpora, the token- and type-level frequency distributions over major word categories\footnote{We preprocess all corpora using the SpaCy 2.3.2 preprocessing pipeline with the \texttt{en\_core\_web\_lg} model. For our data and Brent, we process the entire corpus. Since MSR, R2R, and Wikipedia are much larger, we process a random sample of 5K sentences from each.} and of individual lexical items.
In terms of word categories, we see that our data most closely mirrors the distribution of child-directed speech: Both our corpus and the Brent corpus contain primarily verbs ($\sim$23\% when computed at the token level) followed by pronouns ($\sim$19\%) followed by nouns at around 17\%. In contrast, the MSR video caption corpus and Wikipedia both contain predominantly nouns ($\sim$40\%) and the R2R instruction dataset contains nouns and verbs in equal proportions ($\sim$33\% each). None of the baseline corpora contain significant counts of pronouns. Additionally, in terms of specific vocabulary items, our corpus contains decent coverage for many of the most frequent verbs observed in CDS, while the baseline corpora are dominated by a single verb each (\textit{``go''} for R2R and \textit{``be''} for MSR and Wikipedia). For nouns and adjectives, we also see better coverage of top-CDS words in our data compared to the other corpora analyzed, though we note that the difference is less obvious and that the lexical items in these categories are much more topically determined.  

\subsubsection{Word-Context Alignment}
\label{sec:cps:signal}

We next look at how well the language corresponds to the the salient objects and events in the context of its use. This property is important as it relates to how strong the ``training signal'' would be for a model that is attempting to learn linguistic meaning from distributional signal. It is hard to directly estimate the quality of the ``training signal'' available to children. However, experiments in psychology using the Human Simulation Paradigm (HSP) \cite{gillette1999human,piccin2007nouns} come close. In the HSP design, experimenters collect audio and video recordings of a child's normal activities (i.e.\ via head-mounted cameras). Given this data, adults are asked to view segments of videos and predict which words are said at given points in time. This technique is used to estimate how ``predictable'' language is given only the grounded (non-linguistic) input to which a child has access. Using this technique, \citet{gillette1999human} estimates that nouns can be predicted at 45\% accuracy and verbs at 15\% accuracy.

While not directly comparable to our setting, this provides us with an approximate point of comparison against which to benchmark the word-to-context alignment of our collected data. Rather than try to guess the word given a video clip, we instead view a short (5 second) video clip alongside an uttered word and make a binary judgement for whether or not the clip depicts an instance of the word: e.g., \textit{yes or no, does the clip depict an instance of ``pick up''?} We chose this design over the HSP design since it provides a more interpretable measure of the quality of the training signal from the perspective of NLP and ML researchers using the data. We expect this variant of the task to yield higher numbers than the HSP design, since it does not require guessing from the entire vocabulary. We take a sample of (up to) five instances for each of our target nouns and verbs (fewer if the word occurs less often in our data) and label them in this way. We find inner annotator agreement on this task to be very high (91\% when computed between two researchers on the project) and thus have a single annotator label all instances. 

Table \ref{tab:train-signal} shows the results of this analysis. We see the expected trend, in which grounded context is a considerably better signal of noun use than verb use. We also note there is substantial variation in training signal across verbs. For example, while some verbs (e.g. \textit{``pick''}, \textit{``take''}, \textit{``hold''}) have strong signal, other verbs (\textit{``eat''}) tend to be used in contexts sufficiently detached from the activities themselves. The noisiness of this signal is one of the biggest challenges of learning from such naturalistic data, as we will discuss further in \S\ref{sec:results}.

\begin{table}[ht!]
\centering
\begin{tabular}{lrclrc}
	\toprule
 \multicolumn{3}{c}{Nouns} & \multicolumn{3}{c}{Verbs}  \\
$w$  & N & P & $w$ & N & P \\
 \cmidrule(lr){1-3}  \cmidrule(lr){4-6}
table	&	81	&	1.0	&	go	&	238	&	0.0	\\
spoon	&	76	&	1.0	&	put	&	193	&	0.4	\\
banana	&	75	&	0.8	&	pick	&	162	&	0.8	\\
apple	&	68	&	1.0	&	eat	&	77	&	0.0	\\
cup	&	57	&	1.0	&	take	&	63	&	0.8	\\
ball	&	54	&	0.6	&	get	&	43	&	0.4	\\
toy	&	48	&	1.0	&	wash	&	38	&	0.6	\\
fork	&	47	&	0.8	&	play	&	37	&	0.8	\\
bowl	&	42	&	1.0	&	walk	&	25	&	0.4	\\
knife	&	40	&	0.8	&	throw	&	25	&	0.6	\\
book	&	25	&	1.0	&	hold	&	21	&	1.0	\\
plant	&	22	&	1.0	&	drop	&	17	&	0.4	\\
bear	&	18	&	1.0	&	stop	&	13	&	0.0	\\
chair	&	16	&	0.4	&	give 	&	13	&	0.0	\\
doll	&	13	&	0.8	&	open	&	3	&	0.3	\\
clock	&	12	&	0.6	&		&		&		\\
lamp	&	2	&	1.0	&		&		&		\\
door	&	2	&	0.0	&		&		&		\\
window	&	1	&	1.0	&		&		&		\\
 \cmidrule(lr){1-3}  \cmidrule(lr){4-6}
 Avg. & 37 &  0.8 & Avg. & 64 & 0.4 \\
\bottomrule
\end{tabular}
\caption{Estimates of training signal quality for nouns and verbs. N is the number of times the word occurs in the training data. P is the precision--given a 5 second clip in which the word is used, how often does the clip depict an instance of the word? Note that the verb \textit{``go''} is an outlier, since it appears most often as \textit{``going to''}.}
\label{tab:train-signal}
\end{table}

\section{Experiments}
\label{sec:exp-design}

Using the above data, we now compare several grounded distributional semantics models (DSM) in terms of how well they encode verb meanings, focusing in particular on differences in how the environment is represented when put in to the DSM. Our hypothesis is that models will perform better if they represent the environment in terms of 3D objects and their physics rather than pixels, since work in psychology has shown that children learn to parse the physical world into objects and agents very early in life \cite{spelke2007core}, long before they show evidence of language understanding. We also explore how models vary when they have access to linguistic supervision early in the pipeline, during environment encoding, in addition to later, during language learning. We note that the models explored are intended as simple instantiations to test the parameters of interest given our (small) dataset. Future work on more advanced models should no doubt yield improvements. 

\subsection{Preprocessing}
\label{sec:preproc}

Our raw data consists of continuous video and game-engine recordings of the environment, and parallel transcriptions of the natural language narration. To convert this into a format usable by our DSM, we perform the following preprocessing steps. This preprocessing phase is common to all the models evaluated. First, we segment the environment data into ``clips''. Each clip is five seconds long\footnote{The length of 5 seconds was chosen heuristically prior to model development.} and thus consists of 450 frames (since the VR environment recording is at 90fps), which we subsample to 50 frames (10fps). Since our grounded DSMs require associating a word $w$ with its grounded context $c$, we consider the clip immediately following the utterance of $w$ to be the context $c$. See earlier discussion (\S\ref{sec:cps:signal}) for estimates of the signal-to-noise ratio produced by this labeling method. Training clips that are not the context of any word are discarded.
We hold out two subjects' sessions (one from each visual aesthetic) for test, and use the remaining 16 subjects' sessions for training.

Finally, since this verb-learning problem proves quite challenging, we scope down our analysis to the following 14 verbs, which come from the 20 verbs specified in our initial target vocabulary (\S\ref{sec:protocol}) less 6 which did not ultimately occur in our data: \textit{``walk'', ``throw'', ``put (down)'', ``get'', ``go'', ``give'', ``wash'', ``open'', ``hold'', ``eat'', ``play'', ``take'', ``drop'', ``pick (up)''}. Again, these words all have low average ages of acquisition (19 to 28 months) and thus should represent reasonable targets for evaluation. Nonetheless, we will see in \S\ref{sec:eval} that models struggle to perform well on this task; we elaborate on this discussion in \S\ref{sec:discussion}. 
 
\subsection{Models}

We train and evaluate four different DSMs, each of which represent a word $w$ in terms of its grounded context $c$. The parameters we vary are 1) the feature representation of $c$ (\$\ref{sec:encoders}) and 2) the type of supervision provided to the DSM (\S\ref{sec:dimred}). All models share the same simple pipeline. First, we build a word-context matrix $M$ which maps each token-level instance of $w$ to a featurized representation of $c$. We then run dimensionality reduction on $M$. Finally, we take the type-level representation of $w$ to be the average row vector of $M$, across all instances of $w$. All of our model code is available at \url{http://github.com/dylanebert/nbc_starsem}. 

\subsubsection{Context Encoders}
\label{sec:encoders}

\paragraph{Object-Based.} 

In our Object-Based encoder, we take a feature-engineered approach intended to provide the model with a knowledge of the basic object physics likely to be relevant to the semantics of the verbs we target. Specifically, we represent each clip using four feature templates (\texttt{trajectory}, \texttt{vel},  \texttt{dist\_to\_head},  \texttt{relPos}), defined as follows. First, we find the ``most moving object'', i.e., the object with the highest average velocity over the clip. We then compute our four sets of features for this most moving object. Our \texttt{velocity} and \texttt{relPos} features are simply the \texttt{mean, min, max, start, end}, and \texttt{variance} of the object's velocity and relative position, respectively, over the clip. For our \texttt{dist\_to\_head} feature, for each position dimension (xyz), we compute the following values of the distance from the object's center to the participant's head: \texttt{start, end, mean, var, min, max, min\_idx, max\_idx}, where min/max index is the point at which min/max value was reached (recorded as a \% of the way through the clip). Finally, our \texttt{trajectory} features are intended to capture the shape of the objects trajectory over the clip. To compute this, for each of position dimension (xyz), we compute four points during the clip: start, peak (max), trough (min), end. Then, if max happens before min, we consider the max to be ``key point 1'' (kp1) and the min to be ``key point 2'' (kp2), and vice-versa if the min happens before the max. We then compute the following features: \texttt{kp1-start, kp2-kp1, end-kp2, end-start}.  

\paragraph{Pretrained CNN.}

To contrast with the above featured-engineered approach, we also implement an encoder based on the features extracted by a pretained CNN. Our CNN encoder has an advantage over the Object-Based encoder in that it has been trained on far more image data, but has a disadvantage in that it lacks domain-specific feature engineering. We use pretrained VGG16 \cite{simonyan2014very}, which is a 16-layer CNN trained on ImageNet that produces a 4096-dimensional vector for each image. We compute this vector for each frame in the clip, and then compute the following features along each dimension in order to get a vector representation of the full clip: \texttt{start\_value, end\_value, min, max, mean}. 

\subsubsection{Dimensionality Reduction}
\label{sec:dimred}

Given a matrix $M$ that maps each word instance to a feature vector using one of the encoders above, we run dimensionality reduction to get a 10d vector\footnote{10d is chosen since we are only attempting to differentiate between 14 words, and thus our supervised LDA cannot use more than 13d.} for each word instance. We consider two settings. In the unsupervised setting, we run vanilla SVD. In the supervised setting, we run supervised LDA in which the ``labels'' are the words uttered at the start of the clip as described in \S\ref{sec:preproc}. 

\subsection{Evaluation}
\label{sec:eval}

We evaluate our models in terms of their precision when assigning verbs to unseen clips. Specifically, for our two heldout subjects, we partition the full session into consecutive 5-second clips, resulting in 189 clips total. For testing, unlike in training, we include all clips, even those in which the subject is not speaking. Then, for each model, we encode each clip using the model's encoder and then find the verb with the highest cosine similarity to the encoded clip. The authors then view each clip alongside the predicted verb and make a binary judgement for whether or not the verb accurately depicts the action in the clip, e.g. \textit{yes or no, does the clip depict an instance of ``pick up''?} To avoid annotation bias, all four models plus a random baseline are shuffled and evaluated together, and annotators do not know which prediction comes from which model. Annotator agreement was high (91\%). 

\subsection{Results and Analysis}
\label{sec:results}

\begin{figure*}
    \centering
    \includegraphics[width=\textwidth]{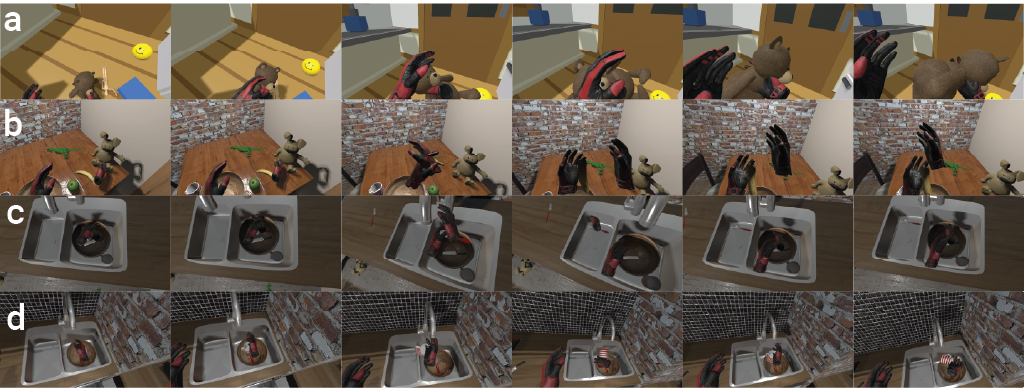}
    \caption{Example clips, subsampled to 6 frames. \textbf{(b)} is \textbf{(a)}'s nearest-neighbor using the Object-Based model. In each of these clips, the participant picks up an object with their right hand. \textbf{(d)} is \textbf{(c)'s} nearest-neighbor using the CNN. In each, the participant is washing dishes in a similar looking sink.}
    \label{fig:sequences}
\end{figure*}

Table \ref{tab:results} reports our main results for each model. 
We compute both ``strict'' precision, in which a prediction is only considered correct if both annotators deemed it correct, as well as ``soft'' precision, in which a prediction is correct as long as one annotator deemed it correct. As the results show, no model performs especially well. Random guessing achieves 32\% (soft) precision on average. The supervised Object-Based model and the unsupervised CNN model both perform a bit better (40\% on average), but we note that the samples are small and we cannot call these differences significant (see 95\% bootstrapped confidence intervals given in Table \ref{tab:results}).  Only the unsupervised Object-Based model stands out in that it performs significantly worse than all other models (20\% soft precision). For the CNN models, we do not see a significant difference with the supervised dimensionality reduction. Figure \ref{fig:sequences} shows example clips for each encoder. 

\begin{table}[ht!]
\centering
\begin{tabular}{lrr}
	\toprule
		& \multicolumn{1}{c}{Soft} &  \multicolumn{1}{c}{Strict} \\
\midrule
Random &$0.32$ ($0.25$--$0.39$) &$0.23$  ($0.17$--$0.29$)\\
\midrule
Obj. &$0.20$  ($0.14$--$0.25$) &$0.13$  ($0.08$--$0.19$)\\
CNN &$0.40$  ($0.33$--$0.47$) &$0.29$  ($0.22$--$0.36$)\\
\midrule
Obj$+$Sup.	&$0.40$  ($0.33$--$0.47$) &$0.28$ ($0.22$--$0.34$)\\
CNN$+$Sup.	&$0.35$  ($0.28$--$0.42$) &$0.25$  ($0.19$--$0.31$)\\
\bottomrule
\end{tabular}
\caption{Precision of each method with 95\% bootstrapped CI. ``Soft'' means a prediction is correct as long as one annotator considers it to be so; ``strict'' means prediction is only considered correct if both annotators agree that it is correct.}
\label{tab:results}
\end{table}

Table \ref{tab:verb-breakdown} shows a breakdown of model performance by verb. 
We see a few intuitive differences between the CNN-based model and the Object-Based model, discussed below. We note these observations are based on a small number of predictions, and thus should be taken only as suggestive. 

\begin{table}[ht!]
\centering
\begin{tabular}{l|rc|rc|rc}
\toprule
& \multicolumn{2}{c|}{CNN}  & \multicolumn{2}{c|}{Obj} & \multicolumn{2}{c}{Obj+Sup.} \\
& N & Prec. & N & Prec. & N & Prec. \\
\midrule
pick & 0 & 0.00 & 1 & 1.00 & 4 & 1.00\\
take & 0 & 0.00 & 3 & 1.00 & 12 & 0.67\\
hold & 11 & 0.64 & 5 & 0.80 & 17 & 0.65\\
get & 32 & 0.56 & 5 & 0.00 & 13 & 0.54\\
go & 29 & 0.21 & 1 & 0.00 & 17 & 0.47\\
put & 4 & 0.00 & 11 & 0.27 & 31 & 0.29\\
play & 7 & 0.29 & 17 & 0.18 & 6 & 0.17\\
walk & 16 & 0.44 & 33 & 0.30 & 26 & 0.15\\
throw & 36 & 0.08 & 25 & 0.00 & 16 & 0.06\\
drop & 4 & 0.25 & 16 & 0.06 & 2 & 0.00\\
eat & 2 & 0.00 & 2 & 0.00 & 19 & 0.00\\
give & 17 & 0.00 & 35 & 0.00 & 5 & 0.00\\
open & 8 & 0.00 & 30 & 0.00 & 10 & 0.00\\
wash & 23 & 0.48 & 5 & 0.00 & 11 & 0.00\\
\bottomrule
\end{tabular}
\caption{Analysis of model precision broken down by verb. Top-level columns are the unsupervised CNN, unsupervised obj model, and supervised obj model.\footnote{Supervised CNN is excluded because it's similar to Unsupervised CNN} For each, N is the number of times the model predicts that verb. Precision is the proportion of the time that prediction was correct.}
\label{tab:verb-breakdown}
\end{table}

\paragraph{Low-level actions.}
The Object-Based models achieve higher precision on low-level verbs like \textit{``pick''}, \textit{``take''}, and \textit{``hold''}. This makes intuitive sense, since the 3D spatial features are designed to capture these types of mechanical actions, independent of the objects with which they co-occur. The 2D visual data, on the other hand, may struggle to ground a visually diverse set of objects-in-motion to these low-level mechanical actions.

\paragraph{Visual cues.}
Some actions are strongly predicted by specific objects, which are well captured by visual cues. This is most obvious in the case of \textit{``wash''}, on which the CNN achieves higher precision than the Object-Based models. This is again intuitive as \textit{wash} tends to co-occur with a clear view of the sink, which is a large, visually-distinct part of the field of view.

\paragraph{Vague actions.}
Actions like \textit{``go''}, \textit{``walk''}, and \textit{``hold''} occur frequently, even when the language signal does not reflect it. That is, in any given clip, there is a high chance that the participant walks, goes somewhere, or holds something. Thus, models which happen to predict these verbs frequently may have artificially high accuracy. For example, the unsupervised Object-Based model only predicts \textit{``go''} once and \textit{``hold''} 5 times
, which may contribute to the unsupervised Object-Based model performing significantly worse than random, despite seeming to capture low-level actions well.

\paragraph{Special cases.}
We note that some verbs are very difficult or impossible to detect given limitations of our data. In particular, \textit{``give''}, \textit{``eat''}, and \textit{``open''} have a precision of 0 across all models, as well as in the training signal (\S\ref{sec:cps:signal}). For example, \textit{``give''} only occurs twice in our data (\textit{``fluffy teddy bear going to give it a little hug''} and \textit{``turn on the water give it a little sore[sic] and we can let it dry there''}), but cannot occur in its prototypical sense since there is no clear second agent to be a recipient. During instances of \textit{``eat''} and \textit{``open''}, participants tended to mime the actions, but the in-game physics data does not faithfully capture the semantics of these verbs (e.g., containers do not actually open). These words highlight limitations of the environment which may be addressed in future work. 


\section{Discussion}
\label{sec:discussion}

We compare two types of models for grounded verb learning, one based on 2D visual features and one based on 3D symbolic and spatial features. Our analysis suggests that these approaches favor in different aspects of verb semantics. One open question is how to combine these differing signals, and how to design training objectives that encourage models to chose the right sensory inputs and time scale to which to ground each verb. 

We evaluated on a small set of verbs that are acquired comparably early by children. Nonetheless, our models perform only marginally better than random. This disconnect highlights an important challenge to be addressed by work on computational models of grounded language learning: Can statistical associations between words and contexts result in more than simple noun-centric image or video captioning, eventually forming general-purpose language models? While that question is still wide open, research from psychology could better inform work on grounded NLP.
For example, \citet{piccin2007nouns} argues that verb learning in particular is not learned from purely grounded signal, but rather is ``scaffolded'' by earlier-acquired knowledge of nouns and of syntax. From this perspective, the models we explored here, which are similar to what is used for noun-learning, are far too simplistic for verb learning. More research is needed on ways to combine linguistic and grounded signal in order to learn more abstract semantic concepts.    




\section{Related Work}

We contribute to a large body of research on learning grounded representations of language. Grounded representations have been shown to improve performance on intrinsic semantic similary metrics \cite{hill2017understanding,vulic2017hyperlex} as well as to be better predictors of human brain activity \citep{anderson2015reading,bulat-clark-shutova:2017:EMNLP2017}.  Much prior work has explored the augmentation of standard language modeling objectives with 2D image \citep{bruni-etal-2011-distributional,kiela2017learning,lazaridou2015combining,silberer-lapata:2012:EMNLP-CoNLL,divvala2014learning} and video \citep{sun2019videobert} data. Recent work on detecting fine-grained events in videos is particularly relevant \citep[among others]{hendricks-etal-2018-localizing,zhukov2019cross,fried-etal-2020-learning}. Especially relevant is the data collected by \citet{gaspers-etal-2014-multimodal}, in which human subjects were asked to play simple games with a physical robot and narrate while doing so. Our data and work differs primarily in that we focus on the ability to ground to symbolic objects and physics rather than only to pixel data.  
Past work on ``situated language learning'', inspired by emergence theories of language acquisition \cite{macwhinney2013emergence}, has trained AI agents to learn language from scratch by interacting with humans and/or each other in simulated environments or games \citep{wang-etal-2016-learning-language,mirowski2016learning,urbanek-etal-2019-learning,beattie2016deepmind,hill2018understanding,mirowski2016learning},

\section{Conclusion}

We introduce the New Brown Corpus, a dataset of spontaneous speech aligned with rich environment data, collected in a VR kitchen environment. We show that, compared to existing corpora, the distribution of vocabulary collected is more comparable to that found in child-directed speech. We analyze several baseline distributional models for verb learning. Our results highlight the challenges of learning from naturalistic data, and outlines directions for future research.

\section{Acknowledgements}

This work was supported by DARPA under award number HR00111990064. Thanks to George Konidaris, Roman Feiman, Mike Hughes, members of the Language Understanding and Representation (LUNAR) Lab at Brown, and the reviewers for their help and feedback on this work. 

\bibliography{nbc}
\bibliographystyle{acl_natbib}

\clearpage






\begin{appendix}

\section{Target Vocabulary}
\label{app:vocab}

\begin{table}[ht!]
\small
\centering 
\begin{tabular}{p{.95\linewidth}}
	\toprule
\multicolumn{1}{l}{\bf Target Vocabulary} \\
    \midrule
{\bf Nouns}: ball (16), banana (16), book (16), shoe (16), apple (18), bear (18), cup (19), spoon (19), door (20), toy (21), chair (21), doll (22), fork (22), bowl (22), clock (23), table (23), knife (25), window (25), plant (26), lamp (28)\\
{\bf Verbs}: eat (19), go (19) play (22), open (22), stop (22), walk (22), get (23), wash (23), cook (24), close (24), push (24), throw (24), hold (25),  drop (26), give (26), put (26), take (26), pick (28), shake (28) \\
{\bf Prep./Adj.}: up (18), down (19), outside (19), on (22), red (23), bad (24), green (24), good (24), in/inside (24), yellow (24), fast (25), here (25), back (26) full (26), there (26), empty (27), away (28), by (28), slow (28), to (28) \\
    \bottomrule
\end{tabular}
\caption{Target vocabulary items and median age of acquisition in months (i.e.\ age at which 50\% of children were reported to have known the word according to \citet{frank2017wordbank}.}
\label{tab:target-vocab}
\end{table}

\section{Assignment of Participants to Test Conditions}
\label{app:conditions}

In order for the collected data to serve as a useful testbed for computational models, it needs to support evaluation of language representations outside of the train environment. Thus, we attempt to distribute vocabulary usage uniformly across our six different environments. To a large extent, we hope that this will occur naturally by virtue of the fact that every participant performs the same tasks, and thus is likely to refer to the same objects/actions. However, to control this more directly, we distribute ``focus objects'' (that is, the objects participants are asked to describe in Task 5) evenly across kitchens. Specifically, during Task 5, we ask each participant to describe 4 distinct objects, one at a time. We divide 12 target objects across 18 test conditions, so that every object is given as a focus object exactly once in each kitchen. This ensures that each of the focus nouns (and, ideally, its co-occurring verbs and modifiers) will be discussed at least once in each kitchen. Table \ref{tab:conditions} shows how our 18 test conditions were constructed in order to distribute focus objects uniformly across kitchens.


\begin{table*}[ht!]
\begin{tabular}{cccccccc}
	\toprule
	&	Participant	&	Aesthetic	&	Layout	&	\multicolumn{4}{c}{Focus Objects}\\
\midrule
1	&	1 1a	&	Blocky	&	1	&	fork	&	banana	&	book	&	bear/bunny	\\
2	&	2 2a	&	PhotoReal	&	1	&	fork	&	banana	&	book	&	bear/bunny	\\
3	&	3 1b	&	Blocky	&	2	&	plant	&	spoon	&	cup	&	truck/plane	\\
4	&	4 2b	&	PhotoReal	&	2	&	plant	&	spoon	&	cup	&	truck/plane	\\
5	&	5 1c	&	Blocky	&	3	&	apple	&	ball	&	doll/dinosaur	&	bowl	\\
6	&	6 2c	&	PhotoReal	&	3	&	apple	&	ball	&	doll/dinosaur	&	bowl	\\
7	&	7 1a	&	Blocky	&	1	&	apple	&	ball	&	cup	&	spoon	\\
8	&	8 2a	&	PhotoReal	&	1	&	apple	&	ball	&	cup	&	spoon	\\
9	&	9 1b	&	Blocky	&	2	&	book	&	banana	&	bowl	&	doll/dinosaur	\\
10	&	10 2b	&	PhotoReal	&	2	&	book	&	banana	&	bowl	&	doll/dinosaur	\\
11	&	11 1c	&	Blocky	&	3	&	fork	&	bear/bunny	&	truck/plane	&	plant	\\
12	&	12 2c	&	PhotoReal	&	3	&	fork	&	bear/bunny	&	truck/plane	&	plant	\\
13	&	13 1a	&	Blocky	&	1	&	truck/plane	&	plant	&	doll/dinosaur	&	bowl	\\
14	&	14 2a	&	PhotoReal	&	1	&	truck/plane	&	plant	&	doll/dinosaur	&	bowl	\\
15	&	15 1b	&	Blocky	&	2	&	apple	&	ball	&	bear/bunny	&	fork	\\
16	&	16 2b	&	PhotoReal	&	2	&	apple	&	ball	&	bear/bunny	&	fork	\\
17	&	17 1c	&	Blocky	&	3	&	banana	&	book	&	cup	&	spoon	\\
18	&	18 2c	&	PhotoReal	&	3	&	banana	&	book	&	cup	&	spoon	\\
\bottomrule
\end{tabular}
\caption{18 test conditions used to distribute vocabulary across kitchen environments.}
\label{tab:conditions}
\end{table*}

\section{Verbal Instructions Given to Participants}
\label{app:script}

\subsection{Intro Instructions}

[Say] The point of this project is to collect data to train a model to learn language from speech that is similar to what a young child might hear. This research is funded by the U.S.\ Department of Defense. We’re going to divide the session into 10 short segments, each about 1 or 2 minutes long. I will give you instructions at the start of each segment. We will record video of the virtual environment and audio of your speech. The data will be released for research purposes, and your audio will be included in that data release. Your name will not appear anywhere. Can we ask you to read and sign the consent form acknowledging that you agree to allow us to use and/or release each component of the recording?

[Walk them through and have them sign the consent form.]

\subsection{Warm up/Getting Acquainted}

[Say] Okay, to start we will ask that you walk around and get comfortable with the environment. During this segment, try to interact with everything so that you understand how the objects behave and are not surprised during the real recording. For example, you will notice that the sink will not produce water, the trash can does not open, etc. We will record video but no audio during this segment, so feel free to ask me questions or comment on the environment however you would like to. To avoid running into objects in the real world, try to stay inside the blue boundaries. Take your time, and just let me know when you feel comfortable and want to move on to the next segment. Do you have questions before we begin?

[Say] Okay, I am starting the video recording now.

[Start new video]

[Record until task complete]

[Stop recording]

[Say] Okay, you are no longer being recorded.

[Save video] Save video/metadata under name: subjectId\_sceneID\_warmup

\subsection{Overall Instructions}

[Say] I am going to reset the environment to its initial state now.

[Reset environment] 

[Say] Okay, now we will begin the language recordings. The following instructions pertain to the rest of the session. I will give you other additional instructions as we go along.

[Say] Throughout all segments, imagine that you are a babysitter or parent speaking to a pre-verbal toddler (e.g.\ less than a year old) and try to speak as naturally as possible given that context. Try to use short sentences with simple words and descriptions. For example, use sentences like  “This is an apple.” “Look at the apple.” “I am picking up the apple.”

[Say] Try to align your language with the actions you are doing in real time. For example, as you are picking up dishes you might say “Okay let’s pick up the fork and knife now. Now let’s bring them over to the table.” It is also okay to comment on what you are going to do before you do it. For example, you can say “I am looking for a bowl. Oh, there it is. Okay I am picking up the bowl.”

[Say] It is okay (and encouraged) to make generalizations and to use prescriptive language when its natural. For example, while you are carrying the knife, its okay to say “See how slowly I am walking? I want to be careful when I carry the knife because the knife is sharp!” Or to say “Okay now let’s bring the apple over to the table. Look how red and shiny the apple is! Doesn’t it look good and healthy?”

[Say] It is also okay (and encouraged) to comment on things that don’t go as planned. For example, if you drop the fork you can say “Oh no! I dropped the fork. I am picking it up off the floor now.”

[Say] The only thing you should not comment on is the notion of the VR environment itself or on ways in which its appearance or physics are mismatched with reality. For example, don’t say things like “This spoon looks kind of weird” or “Usually there is water in the sink but there isn’t water here, so just pretend there is.”

[Say]  Do you have any general questions before we go on to the first segment?

\subsection{Task 1: Set table for meal}

[Say] First, set up the table for lunch. The meal will be soup (you will need to pretend there is soup in the bowl) and fruit (both the banana and the apple). Right now, just set the table with the bowl, all of the utensils, both pieces of fruit, and a glass of water. Don’t pretend to eat anything yet. Remember, try to continue talking throughout the entire session. When you are done, stop interacting, pause, and say something like “All done!” or “All set!” Do you have questions before we begin?

[Say] Okay, I am starting the video recording now.

[Start new video]

[Record until task complete]

[Stop recording]

[Say] Okay, you are no longer being recorded.

[Save video] Save video/metadata under name: subjectId\_sceneID\_task1 

\subsection{Task 2: Eat soup and fruit}

[Say] Okay, now you will eat the lunch you just set up. Again, when you are done, stop interacting, pause, and say something like “All gone!”. Do you have questions before we begin?

[Say] Okay, I am starting the video recording now.

[Start new video]

[Record until task complete]

[Stop recording]

[Say] Okay, you are no longer being recorded.

[Save video] Save video/metadata under name: subjectId\_sceneID\_task2

\subsection{Task 3: Clean up and wash dishes}

[Say] Now you will clean up. Bring the dishes to the sink and pretend to wash them and then leave them on the counter to dry. Put the fruit on the trash can (pretend as though you are opening it and putting them inside, but you will actually have to leave them on top). When you are done, say something like “All done!” or “All clean!” Do you have questions before we begin?

[Say] Okay, I am starting the video recording now.

[Start new video]

[Record until task complete]

[Stop recording]

[Say] Okay, you are no longer being recorded.

[Save video] Save video/metadata under name: subjectId\_sceneID\_task3 

\subsection{Task 4: Play with toys (unstructured)}

[Say] Okay, now is play time. You can spend a few minutes playing with the toys however you want to, then we’ll ask you do play with a few of the items specifically. In this session, try to involve all the toys and remember to comment on all of your actions. For example, if you are going to pick up the doll, say “I am going to find the doll. There it is. I am picking up the doll now.” That kind of thing. After two minutes or so, I will cut you off, so you don’t need to cue when you are done. Do you have questions before we begin?

[Say] Okay, I am starting the video recording now.

[Start new video]

[Record until task complete] Record for at least two minutes. Then, wait until there is a natural lull in their talking and interrupt them to stop the recording. We’ll crop out the last couple seconds, so that the video ends on a natural utterance. 

[Stop recording]

[Say] Okay, you are no longer being recorded.

[Save video] Save video/metadata under name: subjectId\_sceneID\_task4

\subsection{Task 5a: Play with Object 1}

[Say] I am going to ask you to play with a few toys specifically. The focus is on very “child directed speech” here, so you should talk about the toys like you are trying to teach a child about them. For example, comment on things like color, shape, and function. For example, for the ball, you might say things like “Let’s play with the ball now! Look at the ball! Its so yellow! So pretty and yellow, like the sun!” I know this will feel unnatural, but think about that cute little robot trying to learn. :) I’ll cut you off after about 20 seconds. Do you have questions before we begin?

[Say] Okay, first can you play with the {name of object here}.

[Say] I am starting the video recording now.

[Start new video]

[Record until task complete] Record for at least 20 seconds. Then, wait until there is a natural lull in their talking and interrupt them to stop the recording. We’ll crop out the last couple seconds, so that the video ends on a natural utterance. 

\subsection{Task 5b: Play with Object 2}

[Say] Now can you play with the {name of object here}.

[Record until task complete] Record for at least 20 seconds. Then, wait until there is a natural lull in their talking and interrupt them to stop the recording. We’ll crop out the last couple seconds, so that the video ends on a natural utterance. 

\subsection{Task 5c: Play with Object 3}

[Say] Now can you play with the {name of object here}.

[Record until task complete] Record for at least 20 seconds. Then, wait until there is a natural lull in their talking and interrupt them to stop the recording. We’ll crop out the last couple seconds, so that the video ends on a natural utterance. 

\subsection{Task 5d: Play with Object 4}

[Say] Now can you play with the {name of object here}.

[Record until task complete] Record for at least 20 seconds. Then, wait until there is a natural lull in their talking and interrupt them to stop the recording. We’ll crop out the last couple seconds, so that the video ends on a natural utterance. 

[Stop recording]

[Save video] Save video/metadata under name: subjectId\_sceneID\_task5

\subsection{Task 6: Put toys away}

[Say] Alright, for the last segment, you just need to clean up the kitchen by putting all the toys away. The toys go in a pile next to the clock. The book belongs on the table next to the plant. When you are done, say something like “All done!” or “Look how clean the room is!”. Do you have any questions before we start?

[Say] I am starting the video recording now.

[Start new video]

[Record until task complete] 

[Stop recording]

[Say] Okay, you are no longer being recorded.

[Save video] Save video/metadata under name: subjectId\_sceneID\_task6

\section{Data Collection Consent Forms}
\label{app:consent}

\paragraph{Purpose} The goal of this project is to see if a neural network model can learn the meanings of words based on the way they are used in context, from language similar to what  a young child would be likely to hear.

\paragraph{Principle Investigator} The lead researcher on this project is XXX

\paragraph{Funding} This research is funded by XXX

\paragraph{Procedures} We will ask you to carry out simple tasks within a Virtual Reality (VR)  environment. These tasks will include activities like setting a table or playing with a toy. We will ask you to narrate your actions as you do them. If you consent, we will collect audio and video recording of your activities. The following information will be recorded:
\begin{itemize}
\item The state of the environment (e.g.\ locations and velocities of objects)
\item Video recording of the environment (The video will look as it appears to you in VR, meaning it will NOT include video of your actual face/body or any identifying physical attributes)
\item Audio recording of your voice
\end{itemize}
\noindent The collected data will be made available to the research community. You will be associated with your recordings only through a unique numeric ID. None of your personal information (e.g.\ name, title, etc.) will be associated with the released data. 

\paragraph{Time Involved} The recording should take approximately 20 minutes.

\paragraph{Risks} You may feel dizziness or nausea while using the VR headset. If you are uncomfortable and wish to stop, you may end the session at any time.

\paragraph{Benefits} There may be no direct benefits to you for participating in this study. We are not offering any compensation for your involvement in this study.

\paragraph{Confidentiality} You will be associated with your recordings only through a unique numeric ID. None of your personal information (e.g.\ name, title, etc.) will be associated with the released data in any way. 

\paragraph{Voluntary} Participation in this study is voluntary. You will not be penalized in any way if you choose not to participate. Your grades, academic standing, or ability to participate in future research projects will not be affected by your decision to participate or not to participate. 

\paragraph{Contact Information} If you have any questions about your participation in this study, please contact XXX at \url{XXX@XXX.edu}. If you have questions about your rights as a research participant, you can contact XXX University’s Human Research Protection Program at XXX or email them at \url{XXX@XXX.edu}.

\paragraph{Consent to Participate}
Your signature below shows that you have read and understood the information in this document, that you agree to volunteer as a research participant for this study, and that you agree to the recording and release of audio and video during the session.

\begin{itemize}
\item[] Yes/No I agree to participate in the study
\item[] Yes/No I consent to the recording of video of the environment during the session
\item[] Yes/No I consent to the recording of my speech during the session
\item[] Yes/No I consent to the public release of the anonymized recordings  
\end{itemize}

\end{appendix}

\end{document}